Bayesian Approach to Neuro-Rough Models for Modelling HIV

Tshilidzi Marwala and Bodie Crossingham

University of the Witwatersrand

Private Bag x3

Wits, 2050

South Africa

e-mail: t.marwala@ee.wits.ac.za

This paper proposes a new neuro-rough model for modelling the risk of HIV from demographic data. The model is formulated using Bayesian framework and trained using Markov Chain Monte Carlo method and Metropolis criterion. When the model was tested to estimate the risk of HIV infection given the demographic data it was found to give the accuracy of 62% as opposed to 58% obtained from a Bayesian formulated rough set model trained using Markov chain Monte Carlo method and 62% obtained from a Bayesian formulated multi-layered perceptron (MLP) model trained using hybrid Monte. The proposed model is able to combine the accuracy of the Bayesian MLP model and the transparency of Bayesian rough set model.

**Introduction**

The role of machine learning is to be able to make predictions given a set of inputs. However, the other role is to extract rules that govern interrelationships within the data. Machine learning tools such as neural networks are quite good at making predictions given input parameters but are not sufficiently transparent to allow the extraction of linguistic rules that govern the predictions they make. Consequently, they are called 'black-box' tools because they do not give a transparent view of the rules that govern the relationships that make predictions possible.

Rough set theory (RST) was introduced by Pawlak (1991) and is a mathematical tool, which deals with vagueness and uncertainty, and is based on set of rules, which are in terms of linguistic variables. Rough sets are of fundamental importance to computational intelligence and cognitive science and are highly applicable to the tasks of machine learning and decision analysis, especially in the analysis of decisions in which there are inconsistencies. As a consequence of the fact that they are rule-based, rough sets are very transparent but they are not as accurate, and most certainly are not universal approximators, as other machine learning tools such as neural networks in their predictions. It can thus be concluded that in machine learning there is always a trade-off between prediction accuracy and transparency. This paper proposes a combined architecture that takes elements from both rough sets and multi-layered perceptron neural networks. It is, therefore, postulated that this architecture will give a balanced view of the data in terms of both the transparency and accuracy they give.

Rough sets are based on lower and upper approximations of decision classes (Inuiguchib and Miyajima, 2006) and are often contrasted to compete with fuzzy set theory (FST), but it in fact complements it. One of the advantages of RST is that it does not require a priori knowledge about the data set, and it is for this reason that statistical methods are not sufficient for determining the relationship in complex cases such as between the demographic variables and their respective outcomes, as is the case for the application under consideration in this paper.

Greco et al. (2006) generalised the original idea of rough sets and introduced variable precision rough set, which is based on the concept of relative and absolute rough membership. The Bayesian framework is a tool that can be used to extend this absolute to relative. Nishino *et. al.* (2006) applied a rough set method to analyse human evaluation data with much ambiguity such as sensory and feeling data and handles totally ambiguous and probabilistic human evaluation data using a probabilistic approximation based on information gains of equivalent classes. Slezak and Ziarko proposes a rough set model which is concerned primarily with algebraic properties of approximately defined sets and extended the basic rough set theory to incorporate probabilistic information.

This paper proposes a new neuro-rough model and extends this to probabilistic domain using Bayesian framework that is trained using Markov Chain Monte Carlo simulation and Metropolis algorithms. In order to achieve this, the rough set membership functions' granulations and the network weights are interpreted probabilistically as will be seen later in the paper. The proposed neuro-rough model is applied to modelling the relationship between demographic properties and the risk of HIV.

**Rough Set Theory**

Rough set theory deals with the approximation of sets that are difficult to describe with the available information (Orhn and Rowland, 2006). It deals predominantly with the classification of imprecise, uncertain or incomplete information. Some concepts that are fundamental to RST theory are given in the next few sections. The data is represented using an information table and an example for the HIV data set for the $i^{th}$ object is given in Table 1:

Table 1: Information Table of the HIV Data

|  | Race | Mothers' Age | Education | Gravidity | Parity | Fathers' Age | HIV Status |
|---|---|---|---|---|---|---|---|
| $Obj^{(1)}$ | 1 | 32 | 1 | 1 | 2 | 34 | 0 |
| $Obj^{(2)}$ | 2 | 27 | 13 | 2 | 1 | 28 | 1 |
| $Obj^{(3)}$ | 2 | 25 | 8 | 2 | 0 | 23 | 1 |
| . | . | . | . | . | . | . | . |
| $Obj^{(i)}$ | 3 | 27 | 4 | 3 | 1 | 22 | 0 |

In the information table, each row represents a new case (or object). Besides HIV status, each of the columns represents the respective case's variables (or condition attributes). The HIV status is the outcome (also called the concept or decision attribute) of each object. The outcome contains either a 1 or 0, and this indicates whether the particular case is infected with HIV or not.

Once the information table is obtained, the data is discretised into partitions as mentioned earlier. An information system can be understood by a pair $\Lambda = (U, A)$, where $U$ and $A$, are finite, non-empty sets called the universe, and the set of attributes, respectively (Deja and Peszek, 2003). For every attribute *a* an element of A, we associate a set $V_a$, of its values, where $V_a$ is called the value set of *a*.

$$a: U \rightarrow V_a \tag{1}$$

Any subset *B* of *A* determines a binary relation *I(B)* on *U*, which is called an indiscernibility relation. The main concept of rough set theory is an indiscernibility relation (indiscernibility meaning indistinguishable from one another). Sets that are indiscernible are called elementary sets, and these are considered the building blocks of RST's knowledge of reality. A union of elementary sets is called a crisp set, while any other sets are referred to as rough or vague. More formally, for given information system $\Lambda$, then for any subset $B \subseteq A$, there is an associated equivalence relation *I(B)* called the *B − indiscernibility* relation and is represented as shown as:

$$(x, y) \in I(B) \; iff \; a(x) = a(y) \tag{2}$$

RST offers a tool to deal with indiscernibility and the way in which it works is, for each concept/decision *X*, the greatest definable set containing *X* and the least definable set containing *X* are computed. These two sets are called the lower and upper approximation, respectively. The sets of cases/objects with the same outcome variable are assembled together. This is done by looking at the "purity" of the particular objects attributes in relation to its outcome. In most cases it is not possible to define cases into crisp sets, in such instances lower and upper approximation sets are defined. The lower approximation is defined as the collection of cases whose equivalence classes are fully contained in the set of cases we want to approximate (Ohrn and Rowland, 2006). The lower approximation of set *X* is denoted $\underline{B}X$ and is mathematically it is represented as:

$$\underline{B}X = \{x \in U : B(x) \subseteq X\} \tag{3}$$

The upper approximation is defined as the collection of cases whose equivalence classes are at least partially contained in the set of cases we want to approximate. The upper approximation of set *X* is denoted $\overline{B}X$ and is mathematically represented as:

$$\overline{B}X = \{x \in U : B(x) \cap X = \emptyset\} \quad (4)$$

It is through these lower and upper approximations that any rough set is defined. Lower and upper approximations are defined differently in the literature, but it follows that a crisp set is only defined for $\overline{B}X = \underline{B}X$. It must be noted that for most cases in RST, reducts are generated to enable us to discard functionally redundant information (Pawlak, 1991).

*Rough Membership Function*

The rough membership function is described; $\eta_A^X : U \rightarrow [0, 1]$ that, when applied to object *x*, quantifies the degree of relative overlap between the set *X* and the indiscernibility set to which *x* belongs. This membership function is a measure of the plausibility of which an object *x* belongs to set *X*. This membership function is defined as:

$$\eta_A^X = \frac{|[X]_B \cap X|}{[X]_B} \quad (5)$$

*Rough Set Accuracy*

The accuracy of rough sets provides a measure of how closely the rough set is approximating the target set. It is defined as the ratio of the number of objects which can be positively placed in *X* to the number of objects that can be possibly be placed in *X*. In other words it is defined as the number of cases in the lower approximation, divided by

the number of cases in the upper approximation (where $0 \leq \alpha_p(X) \leq 1$) and can be written as:

$$\alpha_p(X) = \frac{|BX|}{|\overline{BX}|} \tag{6}$$

*Rough Sets Formulation*

The process of modeling the rough set can be broken down into five stages. The first stage would be to select the data while the second stage involves pre-processing the data to ensure that it is ready for analysis. The second stage involves discretising the data and removing unnecessary data (cleaning the data). If reducts were considered, the third stage would be to use the cleaned data to generate reducts. A reduct is the most concise way in which we can discern object classes (Witlox and Tindermans, 2004). In other words, a reduct is the minimal subset of attributes that enables the same classification of elements of the universe as the whole set of attributes (Pawlak, 1991). To cope with inconsistencies, lower and upper approximations of decision classes are defined (Ohrn, 2006; Deja and Peszek, 2003). Stage four is where the rules are extracted or generated. The rules are normally determined based on condition attributes values (Goh and Law, 2003). Once the rules are extracted, they can be presented in an *if CONDITION(S)-then DECISION* format (Leke, 2007). The final or fifth stage involves testing the newly created rules on a test set to estimate the prediction error of rough set model. The equation representing the mapping between the inputs $x$ to the output $\gamma$ using rough set can be written as:

$$\gamma = f(G_x, N_r, R) \tag{7}$$

where $\gamma$ is the output, $G_x$ is the granulisation of the input space into high, low, medium etc, $N_r$ is the number of rules and $R$ is the rules. So for a given nature of granulisation, the rough set model will be able to give the optimal number and nature of rules and the accuracy of prediction. Therefore, in rough set modeling there is always a trade-off between the degree of granulisation of the input space (which affects the nature and size of rules) and the prediction accuracy of the rough set model.

*Multi-layer Perceptron Model*

The other component of the neuro-rough model is the multi-layered network. This network architecture contains hidden units and output units and has one hidden layer. The bias parameters in the first layer are weights from an extra input having a fixed value of $x_0=1$. The bias parameters in the second layer are weights from an extra hidden unit, with the activation fixed at $z_0=1$. The model is able to take into account the intrinsic dimensionality of the data. The output of the $j^{th}$ hidden unit is obtained by calculating the weighted linear combination of the $d$ input values to give (Bishop, 2006; Marwala, 2001):

$$a_j = \sum_{i=1}^{d} w_{ji}^{(1)} x_i + w_{j0}^{(1)} \tag{7}$$

Here, $w_{ji}^{(1)}$ indicates weight in the first layer, going from input $i$ to hidden unit $j$ while $w_{j0}^{(1)}$ indicates the bias for the hidden unit $j$. The activation of the hidden unit $j$ is obtained by transforming the output $a_j$ in equation 7 into $z_j$, as follows:

$$z_j = f_{inner}(a_j) \tag{8}$$

The $f_{inner}$ function represents the activation function of the inner layer and functions such as hyperbolic tangent function may be used (Bishop, 1996; Marwala, 201; Marwala,

2007ª). The output of the second layer is obtained by transforming the activation of the second hidden layer using the second layer weights. Given the output of the hidden layer $z_j$ in equation 8, the output of unit k may be written as:

$$a_k = \sum_{j=1}^{M} w_{kj}^{(2)} z_j + w_{k0}^{(2)} \tag{9}$$

Similarly, equation 9 may be transformed into the output units by using some activation function as follows:

$$y_k = f_{outer}(a_k) \tag{10}$$

If equations 7, 8, 9 and 10 are combined, it is possible to relate the input $x$ to the output $y$ by a two-layered non-linear mathematical expression that may be written as follows (Bishop, 1995; Haykin, 1995; Hinton, 1987):

$$y_k = f_{outer}\left(\sum_{j=1}^{M} w_{kj}^{(2)} f_{inner}\left(\sum_{i=1}^{d} w_{ji}^{(1)} x_i + w_{j0}^{(1)}\right) + w_{k0}^{(2)}\right) \tag{11}$$

Models of this form can approximate any continuous function to arbitrary accuracy if the number of hidden units *M* is sufficiently large. The MLP may be expanded by considering several layers but it has been demonstrated by the Universal Approximation Theorem (Haykin, 1999) that a two-layered architecture is adequate for the multi-layer perceptron.

*Neuro-Rough Model*

If equations 7 and 11 are combined, it is possible to relate the input $x$ to the output $y$ by a two-layered non-linear mathematical expression that may be written as follows:

$$y_k = f_{outer}\left(\sum_{j=1}^{M} \gamma_{kj}(G_x, R, N_r) f_{inner}\left(\sum_{i=1}^{d} w_{ji}^{(1)} x_i + w_{j0}^{(1)}\right) + w_{k0}^{(2)}\right) \tag{12}$$

The biases in equation 12 may be absorbed into the weights by including extra input variables set permanently to 1 making $x_0 = 1$ and $z_0 = 1$, to give:

$$y_k = f_{outer}\left(\sum_{j=0}^{M} \gamma_k(G_x, R, N_r)_{kj} f_{inner}\left(\sum_{i=0}^{d} w_{ji}^{(1)} x_i\right)\right) \quad (13)$$

The function $f_{outer}(\cdot)$ may be logistic, linear, or sigmoid while $f_{inner}$ is a hyperbolic tangent function. The equation may be represented schematically by Figure 1.

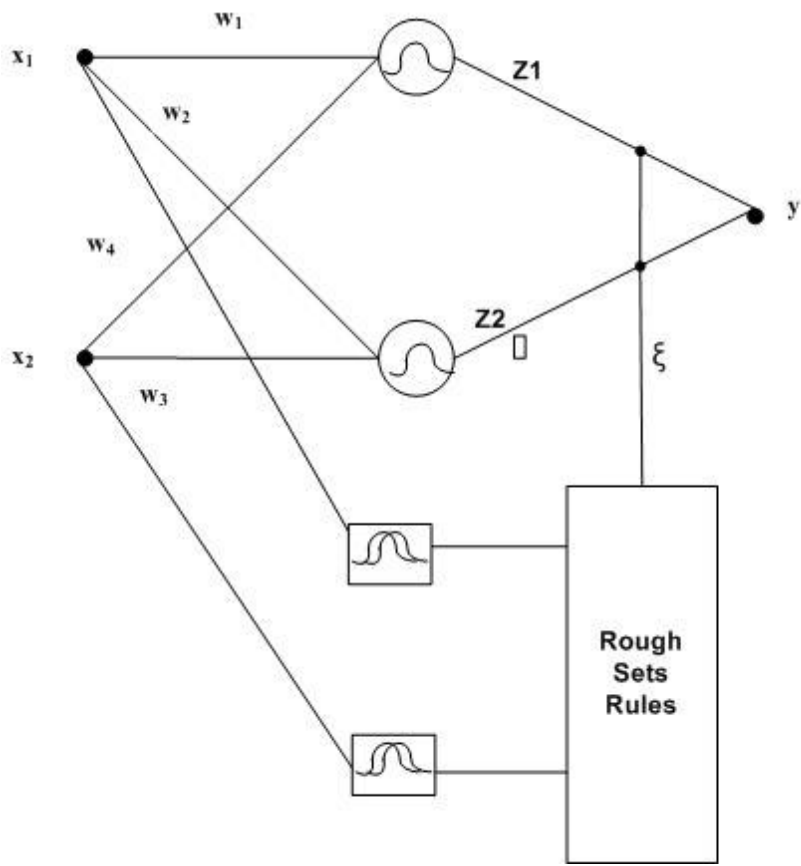

Figure 1: The neuro-rough model

*Bayesian training on rough sets*

The Bayesian framework can be written as in (Marwala, 2007[a,b]; Bishop, 2006):

$$P(M \mid D) = \frac{P(D \mid M) p(M)}{p(D)} \quad \text{where} \quad M = \begin{Bmatrix} w \\ G_x \\ N_r \\ R \end{Bmatrix} \quad (14)$$

The parameter $P(M \mid D)$ is the probability of the rough set model given the observed data, $P(D \mid M)$ is the probability of the data given the assumed rough set model also called the likelihood function, $P(M)$ is the prior probability of the rough set model and $P(D)$ is the probability of the data and is also called the evidence. The evidence can be treated as the normalisation constant. The likelihood function and the resulting error may be estimated as follows:

$$P(D \mid M) = \frac{1}{z_1} \exp(-error) = \frac{1}{z_1} \exp\{A(w, N_r, R, G_x) - 1\} \quad (15)$$

$$error = \sum_l^L \sum_k^K \left( t_{lk} - \left( f_{outer}\left( \sum_{j=0}^M \gamma_{kj}(G_x, R)_{kj} f_{inner}\left( \sum_{i=0}^d w_{ji}^{(1)} x_i \right) \right) \right)_{lk} \right)^2 \quad (16)$$

Here $z_1$ is the normalisation constant, $L$ is the number of outputs while $K$ is the number of training examples. The prior probability in this problem is linked to the concept of reducts, which was explained earlier and it is the prior knowledge that the best rough set model is the one with the minimum number of rules ($N_r$) and that the best network is the one whose weights are of the same order of magnitude. Therefore, the prior probability may be written as follows:

$$P(M) = \frac{1}{z_2} \exp\left\{ -\alpha N_r - \beta \sum w^2 \right\} \quad (17)$$

where $z_2$ is the normalisation constant and $\beta$ is the hyperparameter of the network weights. The posterior probability of the model given the observed data is thus:

$$P(M|D) = \frac{1}{z}\exp\left\{A(N_r, R, G_x) - 1 - \alpha N_r - \beta \sum w^2\right\} \quad (18)$$

where $z$ is the normalisation constant and $\alpha$ is the hyperparameter of the number of rules. Since the number and the rules given the data depends on the nature of granulation, we shall sample in the granule space as well as the network weights using a procedure called Markov Chain Monte Carlo (MCMC) simulation (Marwala, 2007$^a$; Bishop, 2006).

*Markov Monte Carlo Simulation*

The manner in which the probability distribution in equation 18 may be sampled is to randomly generate a succession of granule-weight vectors and accepting or rejecting them based on how probable they are using Metropolis et. al. algorithm (1953). This process requires a generation of large samples of granules for the input space and the network weights, which in many cases is not computationally efficient. The MCMC creates a chain of granules and network weights and accepts or rejects them using Metropolis algorithm. The application of Bayesian approach and MCMC neuro-rough sets, results in the probability distribution function of the granules and network weights, which in turn leads to the distribution of the neuro-rough model outputs. From these distribution functions the average prediction of the neuro-rough set model and the variance of that prediction can be calculated. The probability distributions of neuro-rough set model represented by granules and network weights are mathematically described by equation 18. From equation 18 and by following the rules of probability theory, the distribution of the output parameter, *y*, is written as (Marwala, 2007$^b$):

$$p(y|x,D) = \int p(y|x,M) p(M|D) dM \qquad (19)$$

Equation 19 depends on equation 18, and is difficult to solve analytically due to relatively high dimension of the combined granule and weight space. Thus the integral in equation 19 may be approximated as follows:

$$\tilde{y} \cong \frac{1}{L} \sum_{i=I}^{Z+L-1} F(M_i) \qquad (20)$$

Here *F* is a mathematical model that gives the output given the input, $\tilde{y}$ is the average prediction of the Bayesian neuro-rough set model ($M_i$), *Z* is the number of initial states that are discarded in the hope of reaching a stationary posterior distribution function described in equation 18 and *L* is the number of retained states. In this paper, MCMC method is implemented by sampling a stochastic process consisting of random variables $\{gw_1, gw_2, \ldots, gw_n\}$ through introducing random changes to granule-weight vector $\{gw\}$ and either accepting or rejecting the state according to Metropolis et al. algorithm given the differences in posterior probabilities between two states that are in transition (Metropolis et al., 1953). This algorithm ensures that states with high probability form the majority of the Markov chain and is mathematically represented as:

If $P(M_{n+1}|D) > P(M_n|D)$ then accept $M_{n+1}$, $\qquad (21)$

else accept if $P(M_{n+1}|D) / P(M_n|D) > \xi$ where $\xi \in [0,1]$ $\qquad (22)$

else reject and randomly generate another model $M_{n+1}$.

Basically the steps described above may be summarised as follows:

**Step 1**: Randomly generate the granule weight vector $\{gw\}_n$

**Step 2**: Calculate the posterior probability $p_n$ using equation 18 and vector $\{gw\}_n$

**Step 3**: Introduce random changes to vector $\{gw\}_n$ to form vector $\{gw\}_{n+1}$

**Step 4**: Calculate the posterior probability $p_{n+1}$ using equation 18 and vector $\{gw\}_{n+1}$

**Step 5**: Accept or reject vector $\{gw\}_{n+1}$ using equations 21 and 22

**Step 6**: Go to step 3 and repeat the process until enough samples of distribution in equation 18 have been achieved

**Experimental Investigation: Modelling of HIV**

The proposed method is applied to create a model that uses demographic characteristics to estimate the risk of HIV. In the last 20 years, over 60 million people have been infected with HIV, and of those cases, 95% are in developing countries (Lasry et al, 2007). HIV has been identified as the cause of AIDS. Early studies on HIV/AIDS focused on the individual characteristics and behaviors in determining HIV risk and Fee and Krieger (1993) refer to this as biomedical individualism. But it has been determined that the study of the distribution of health outcomes and their social determinants is of more importance and this is referred to as social epidemiology (Poundstone et. al., 2004). This study uses individual characteristics as well as social and demographic factors in determining the risk of HIV using neuro-rough models formulated using Bayesian approach and trained using Markov Chain Monte Carlo method. Previously, computational intelligence techniques have been used extensively to analyse HIV and Leke et al (2006, 2007) used autoencoder network classifiers, inverse neural networks, as well as conventional feed-forward neural networks to estimate HIV risk from demographic factors. Although good accuracy was achieved when using the autoencoder method, it is disadvantageous due to its "black box" nature which is that it is not transparent. To improve transparency Bayesian rough set theory (RST) was proposed to forecast and interpret the causal effects of HIV (Marwala and Crossingham, 2007) and good accuracy and relevant rules that govern the relationships between demographic characteristics and HIV were identified. Rough sets have been used in various biomedical and engineering applications (Ohrn, 1999; Pe-a et. al, 1999; Tay and Shen, 2003; Golan and Ziarko, 1995). But in most applications, RST is used primarily for prediction. Rowland et. al. (1998) compared the use of RST and neural networks for the prediction of ambulation spinal cord injury, and although the neural network method produced more accurate results, its "black box" nature makes it impractical for the use of rule extraction problems. Poundstone et. al. (2004) related demographic properties to the spread of HIV. In their work they justified the use of demographic properties to create a model to predict HIV from a given database. The data set used in this paper was obtained from the South African antenatal sero-prevalence survey of 2001 (Department of Health, 2001). The data was obtained through questionnaires completed by pregnant women attending selected public clinics and was conducted concurrently across all nine provinces in South Africa. The six demographic variables considered are: race, age of mother, education, gravidity, parity and, age of father, with the outcome or decision being either HIV positive or negative. The HIV status is the decision represented in binary form as either a 0 or 1, with a 0 representing HIV negative and a 1 representing HIV positive. The input data was discretised into four partitions. This number was chosen as it gave a good balance between computational efficiency and accuracy. The parents' ages are given and discretised accordingly, education is given as an integer, where 13 is the highest level of education, indicating tertiary education. Gravidity is defined as the number of times that a woman has been pregnant, whereas parity is defined as the number of times that she has given birth. It must be noted that multiple births during a pregnancy are indicated with a parity of one. Gravidity and parity also provide a good indication of the reproductive health of pregnant women in South Africa. The neuro-rough models were trained by sampling in the granule and weight space and accepting or rejecting samples using Metropolis *et. al.* algorithm (1953).

As with many surveys, there are incomplete entries and such cases are removed from the data set. The second irregularity was information that is false for example an instance where gravidity (number of pregnancies) was zero and parity (number of births) was at least one, which is impossible because for a woman to have given birth she must necessarily have been pregnant. Such cases were removed from the data set. Only 12945 cases remained from a total of 13087. The input data was therefore the demographic characteristics explained earlier and the output were the plausibility of HIV with 1 representing 100% plausibility that a person is HIV positive and -1 indicating 100% plausibility of HIV negative. The neuro-rough model constructed had 7 inputs, 5 hidden nodes, hyperbolic tangent function in the inner layer ($f_{inner}$) and logistic function in the outer layer ($f_{outer}$). When training the neuro-rough models using Markov Chain Monte Carlo, 500 samples were accepted and retained meaning that 500 sets of rules and weights where each set contained 50 up to 550 numbers of rules with an average of 88 rules and the distributions of these rules over the 500 samples are shown in Figure 2. 500 samples were retained because the simulation had converged to a stationary distribution and this can be viewed in Figure 3.

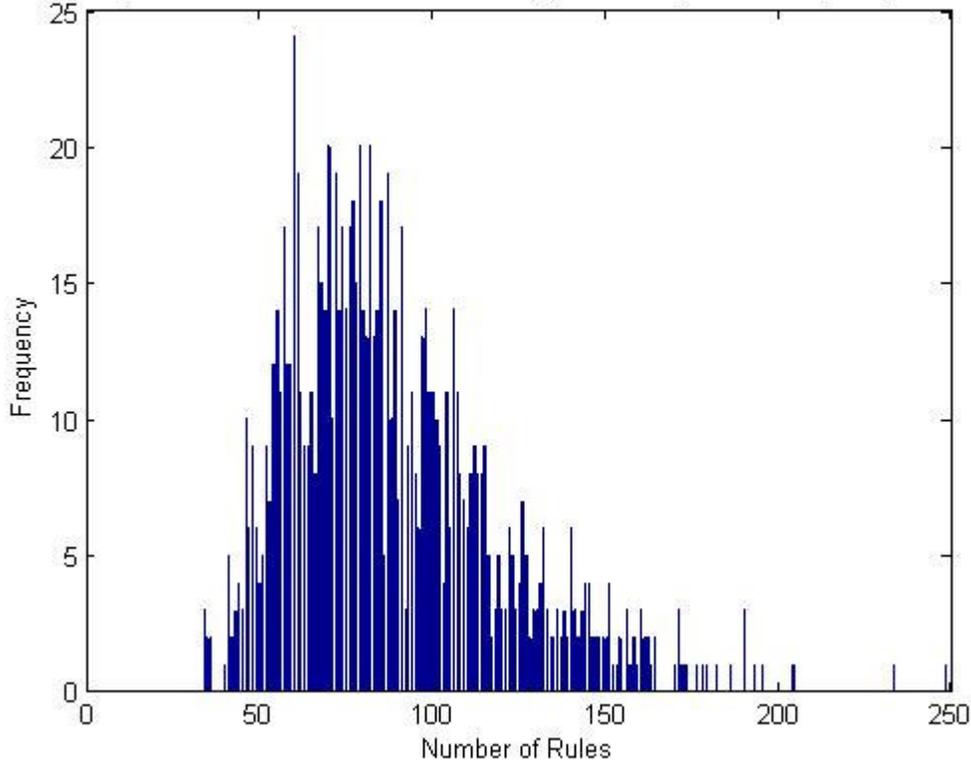

Figure 2: The distribution of number of rules

This figure must be interpreted in the light of the fact that on calculating the posterior probability we used the knowledge that fewer rules and weights of the same order of magnitudes are more desirable. Therefore, the Bayesian neuro-rough model is able to select the number of rules in addition to the partition sizes and weights.

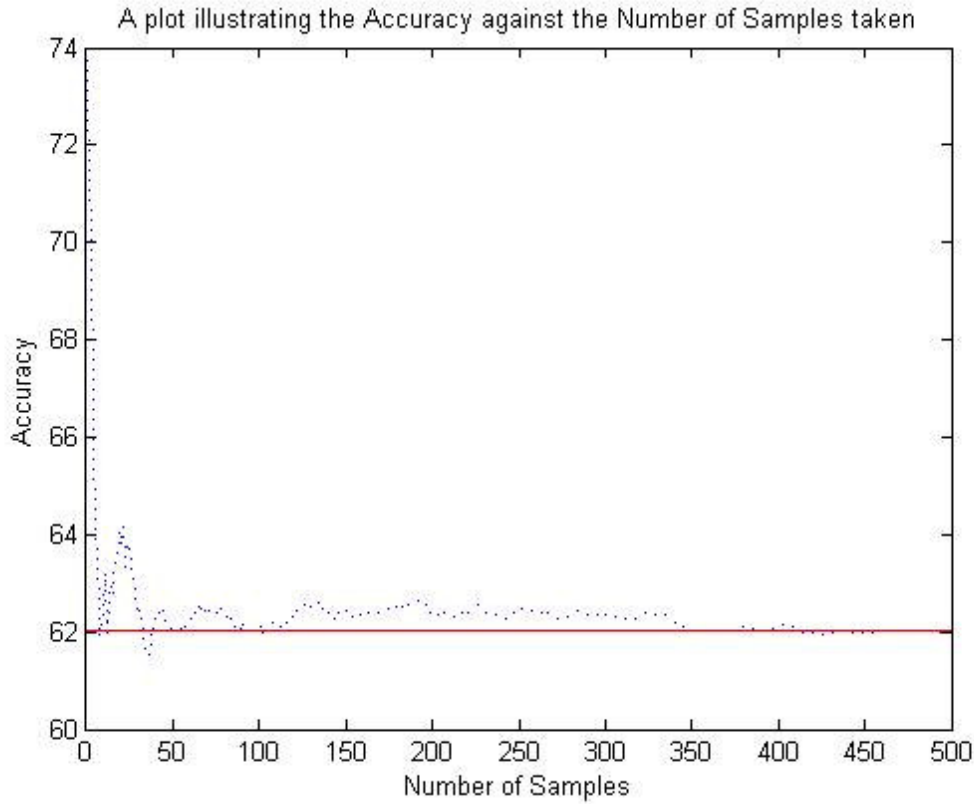

Figure 3: A figure indicating the convergence as the number of samples increase

The average accuracy achieved was 62%. This accuracy can be compared with the accuracy of Bayesian multi-layered perceptron trained using hybrid Monte Carlo by Tim and Marwala (2006), which gave the accuracy of 62%, and Bayesian rough sets by Marwala and Crossingham (2007), which gave the accuracy of 58%, all on the same database. The results show that the incorporation of rough sets into the multi-layered perceptron neural network to form neuro-rough model does not compromise on the results obtained from a traditional Bayesian neural network but it added a dimension of rules. When the interaction between the neural network and the rough set components was conducted by analysing the average magnitudes of the weights ($w$) and the magnitudes of the $\gamma$ which is the output of the rough set model, it was found that the rough set model contributes (average $\gamma$ of 0.43) to the neuro-rough set model less than the neural network component (average $w$ of 0.49). The receiver operator characteristics (ROC) curve was also generated and the area under the curve was calculated to be 0.59 and is shown in Figure 4. This shows that the neuro-rough model proposed is able to estimate the HIV status.

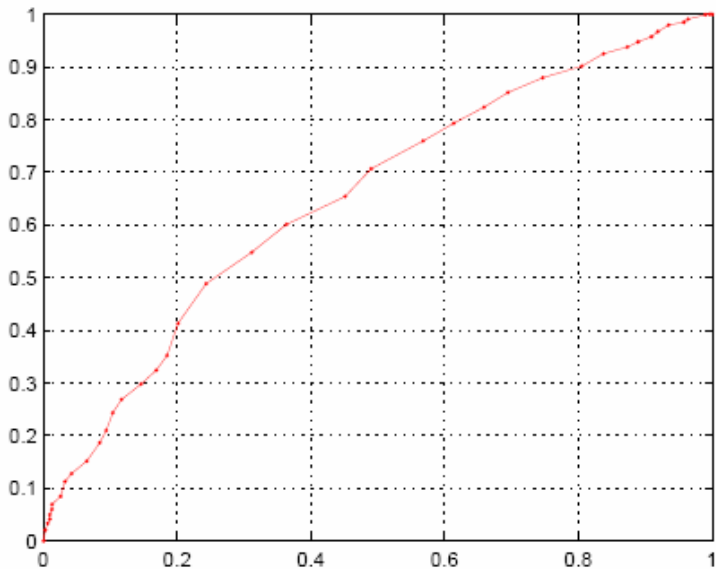

Figure 4: ROC curve of the neuro-rough model with *x*-axis true positive and *y*-axis false positive

Training neuro-rough model with Bayesian framework allows us to determine how confident we are on the HIV status we predict. For example, Figure 5 shows that the average HIV status predicted is 0.8 indicating that a person is probably HIV positive.

The variance of the distribution shown, which is from the 500 samples identified, gives us some measure of the probability distribution of that prediction. This in essence indicates that the Bayesian formulation allows us to interpret the predictions of neuro-rough models in probability terms as can be viewed from a probability distribution.

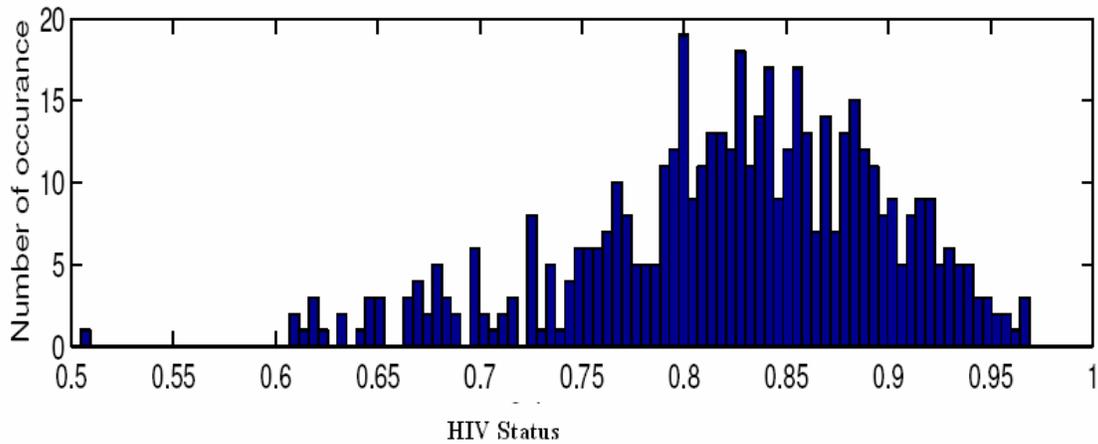

Figure 5: The distribution of the prediction outcome

*Rule Extraction*

Once Bayesian neuro-rough model was applied to the HIV data, unique distinguishable cases and indiscernible cases were extracted. From the data set of 12945 cases, the data is only a representative of 452 cases out of the possible 4096 unique combinations. The lower approximation cases are rules that always hold, or are definite cases while the upper approximation can only be stated with certain plausibility. Examples of both cases that were extracted from the approach in this paper is: **If** Race = AF **and** Mother's Age = Young **and** Education = Secondary **and** Gravidity = Low **and** Parity = Low **and** Father's Age = Young **then** $\gamma = -0.75$ meaning probably HIV negative. This demonstrates that the Bayesian neuro-rough model allows us to extract rules that can be represented in linguistic terms.

**Conclusion**

Neuro-rough model was formulated using Bayesian framework and then trained using Markov Chain Monte Carlo method. The model is able to balance transparency of the rough set model with the accuracy of neural networks. When implemented for HIV estimation it gives 62% accuracy compared to 62% for Bayesian multi-layered networks trained using hybrid Monte Carlo and 59% for Bayesian rough set models.

**Reference**

1. Bishop, C.M. 2006 *Pattern recognition and machine intelligence.* Springer, Berlin, Germany.
2. Department of Health, 2001 *National HIV and syphilis sero-prevalence survey of women attending public antenatal clinics in South Africa*. http://www.info.gov.za/otherdocs/2002/hivsurvey01.pdf.
3. Deja, A. & Peszek, P. 2003 Applying rough set theory to multi stage medical diagnosing. *Fundamenta Informaticae*. 54, 387–408..
4. Fee, E. & Krieger, N. 1993 Understanding AIDS: Historical interpretations and the limits of biomedical individualism. *American Journal of Public Health*, 83, 1477–1486.

5. Greco S., Matarazzo, B. & Slowinski, R. 2006  Rough membership and Bayesian confirmation measures for parameterized rough sets. *Proceedings of SPIE - The International Society for Optical Engineering*, 6104, 314-324.

6. Goh, C. & Law, R. 2003 Incorporating the rough sets theory into travel demand analysis. *Tourism Management* , 24, 511-517.

7. Golan, R.H. & Ziarko, W. 1995 A methodology for stock market analysis utilizing rough set theory. *In Proceedings of computational intelligence for financial engineering*, pp. 32–40.

8. Inuiguchi, M. & Miyajima, T. 2007 Rough set based rule induction from two decision tables. *European Journal of Operational Research*, in press.

9. Lasry, G., Zaric, S. & Carter, M.W. 2007 Multi-level resource allocation for HIV prevention: A model for developing countries. *European Journal of Operational Research*, 180, 786-799.

10. Leke, B.B. 2007 *Computational intelligence for modelling HIV*. Ph.D. thesis, University of the Witwatersrand, School of Electrical and Information Engineering.

11. Leke, B.B., Marwala, T. & Tettey, T. 2006 Autoencoder networks for HIV classification. *Current Science*, 91, 1467–1473.

12. Marwala, T. 2001 *Fault identification using neural networks and vibration data*. Doctor of Philosophy Thesis, University of Cambridge.

13. Marwala, T. 2007[a] Bayesian training of neural network using genetic programming. *Pattern Recognition Letters*,  DOI:http://dx.doi.org/10.1016/j.patrec.2007.03.004 (in press)

14. Marwala, T. 2007[b] *Computational Intelligence for Modelling Complex Systems*. Research India Publishers (in press).

15. Marwala, T & Crossingham, B. 2007 Bayesian approach to rough set. *arXiv* 0704.3433.

16. Leke, B.B., Marwala, T., Tim, T. & Lagazio, M. 2006 Prediction of HIV status from demographic data using neural networks. *In Proceedings of the IEEE International Conference on Systems, Man and Cybernetics. Taiwan*, pp. 2339-2444.

17. Leke, B.B., Marwala & Tettey, T. 2007 Using inverse neural network for HIV adaptive control. *International Journal of Computational Intelligence Research*, 3, 11–15.

18. Metropolis, N., Rosenbluth, A.W., Rosenbluth, M.N., Teller, A.H. & Teller, E., 1953. Equations of state calculations by fast computing machines. *Journal of Chemical Physics*. 21, 1087-1092.

19. Nishino, T., Nagamachi, M. & Tanaka, H. 2006 Variable precision Bayesian rough set model and its application to human evaluation data. *Proceedings of SPIE - The International Society for Optical Engineering*, 6104, 294-303.

20. Ohrn, A. 1999 Discernibility and rough sets in medicine: tools and applications. *PhD Thesis, Department of Computer and Information Science Norwegian University of Science and Technology*.

21. Ohrn, A. & Rowland, T. 2000 Rough sets: A knowledge discovery technique for multifactorial medical outcomes. *American Journal of Physical Medicine and Rehabilitation*, 79, 100-108.

22. Pawlak, Z. 1991 *Rough sets, theoretical aspects of reasoning about data*, Kluwer Academic Publishers, 1991.

23. Pe-a, J., Ltourneau, S., & Famili, A. 1999 Application of rough sets algorithms to prediction of aircraft component failure. *In Proceedings of the third international symposium on intelligent data analysis*. Amsterdam.

24. Poundstone, K.E., Strathdee, S.A. & Celentano, D.D. 2004 The social epidemiology of human immunodeficiency virus/acquired immunodeficiency syndrome. *Epidemiol Reviews*, 26, 22–35.

25. Rowland, T., Ohno-Machado, L. & Ohrn, A. 1998 Comparison of multiple prediction models for ambulation following spinal cord injury. *In Chute*, 31, 528–532.

26. Slezak, D. & Ziarko, W. 2005 The investigation of the Bayesian rough set model. *International Journal of Approximate Reasoning*, 40 (1-2), 81-91

27. Tim, T.N. & Marwala, T. Computational intelligence methods for risk assessment of HIV. *In Imaging the Future Medicine, Proceedings of the IFMBE*, 2006, Vol. 14, pp. 3581-3585, Springer-Verlag, Berlin Heidelberg. Eds. Sun I. Kim and Tae Suk Sah.

28. Tay, F.E.H. & Shen, L. 2003 Fault diagnosis based on rough set theory. *Engineering Applications of Artificial Intelligence*, 16, 39-43.

29. Witlox, F. & Tindemans, H. 2004 The application of rough sets analysis in activity based modelling. Opportunities and constraints. *Expert Systems with Applications*, 27, 585-592.